\documentclass[letterpaper, 10 pt, conference]{ieeeconf}  % Comment this line out if you need a4paper

\IEEEoverridecommandlockouts                              % This command is only needed if 
\usepackage{amssymb} % assumes amsmath package installed

\usepackage{amsthm}

\usepackage{graphicx}
\usepackage{makecell}
\usepackage{threeparttable} % to use table notes
\usepackage{subcaption}
\usepackage{xcolor}
\usepackage{booktabs}
\usepackage{multirow}

\usepackage[pagebackref = false,breaklinks,colorlinks,citecolor=green]{hyperref}

\usepackage{xurl}

\let\titleold\title
\renewcommand{\title}[1]{\titleold{#1}\newcommand{\thetitle}{#1}}

\AtEndPreamble{
    \usepackage[capitalize]{cleveref}
    \crefname{section}{Sec.}{Secs.}
    \Crefname{section}{Section}{Sections}
    \Crefname{table}{Table}{Tables}
    \crefname{table}{Tab.}{Tabs.}
}

\newif\ifproofread
\proofreadfalse
\title{\LARGE \bf
Structureless VIO
}

\author{Junlin Song and Miguel Olivares-Mendez% <-this % stops a space
\thanks{Space Robotics (SpaceR) Research Group, Int. Centre for Security, Reliability and Trust (SnT), University of Luxembourg, Luxembourg.} %{\tt junlin.song@uni.lu} {\tt miguel.olivaresmendez@uni.lu}}
}

\begin{document}

\maketitle
\thispagestyle{empty}
\pagestyle{empty}

%%%%%%%%%%%%%%%%%%%%%%%%%%%%%%%%%%%%%%%%%%%%%%%%%%%%%%%%%%%%%%%%%%%%%%%%%%%%%%%%
\begin{abstract}

Visual odometry (VO) is typically considered as a chicken-and-egg problem, as the localization and mapping modules are tightly-coupled. The estimation of a visual map relies on accurate localization information. Meanwhile, localization requires precise map points to provide motion constraints. This classical design principle is naturally inherited by visual-inertial odometry (VIO). Efficient localization solutions that do not require a map have not been fully investigated. To this end, we propose a novel structureless VIO, where the visual map is removed from the odometry framework. Experimental results demonstrated that, compared to the structure-based VIO baseline, our structureless VIO not only substantially improves computational efficiency but also has advantages in accuracy.

\end{abstract}

% \begin{keywords}
% Visual-Inertial Odometry, Structureless Visual-Inertial Bundle Adjustment
% \end{keywords}

%%%%%%%%%%%%%%%%%%%%%%%%%%%%%%%%%%%%%%%%%%%%%%%%%%%%%%%%%%%%%%%%%%%%%%%%%%%%%%%%

\section{Introduction}
\label{sec:intro}

In the last two decades, the robotics and computer vision communities have designed various VO/SLAM systems that use only a monocular camera \cite{davison2007monoslam, forster2014svo, mur2015orb, engel2017direct}. According to the formulation of visual constraints, these systems can be broadly classified as feature-based methods and direct methods. The common visual constraint used in feature-based methods is the reprojection error whose observed pixel location is from feature tracking or matching. And its predicted pixel location depends on the estimated 3D position of feature point through triangulation with multiple camera poses. The photometric error employed in the direct method typically relies on the depth estimation from coarse to fine. The formulation of aforementioned visual constraints tightly couples the visual map (structure) with localization (pose).

Merely using a monocular camera can not restore the true physical scale due to the scale ambiguity. To address this issue, researchers have proposed to fuse the camera with an additional sensor, like an IMU. Thanks to the built-in accelerometer, IMU can provide scale information to aid visual localization. This fusion scheme is well-known as visual-inertial odometry (VIO). The introduction of IMU significantly increases the output frequency, robustness, and accuracy of odometry; therefore, VIO is widely used in AR/VR \cite{Google, Apple, Meta, fan2024schurvins}, robotics \cite{wu2017vins, delmerico2018benchmark, kang2023view, song2024gps}, and planetary exploration \cite{mourikis2009vision, bayard2019vision, delaune2021range}.

\begin{figure}[htbp]
  \centering
    \begin{subfigure}[t]{0.23\textwidth}
        \centering
        \includegraphics[width=\textwidth, height=\textwidth]{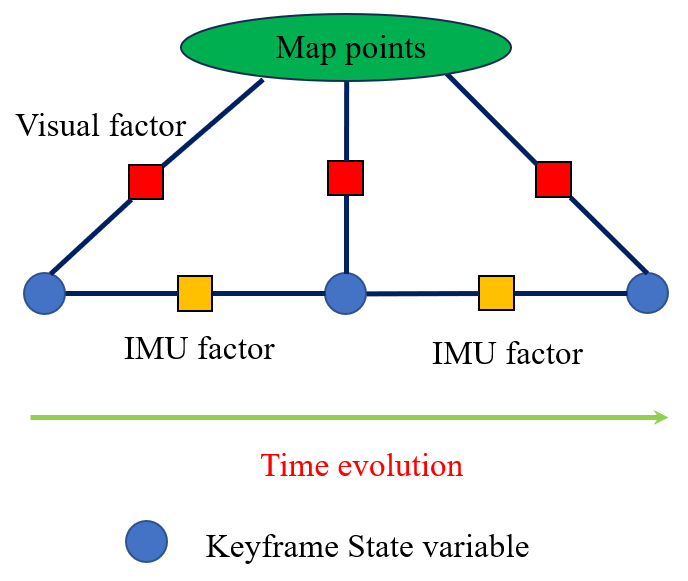}
        % \caption{Structure-based VI-BA}
        \label{factor1}
    \end{subfigure}
    \hfill
    \begin{subfigure}[t]{0.23\textwidth}
        \centering
        \includegraphics[width=\textwidth, height=\textwidth]{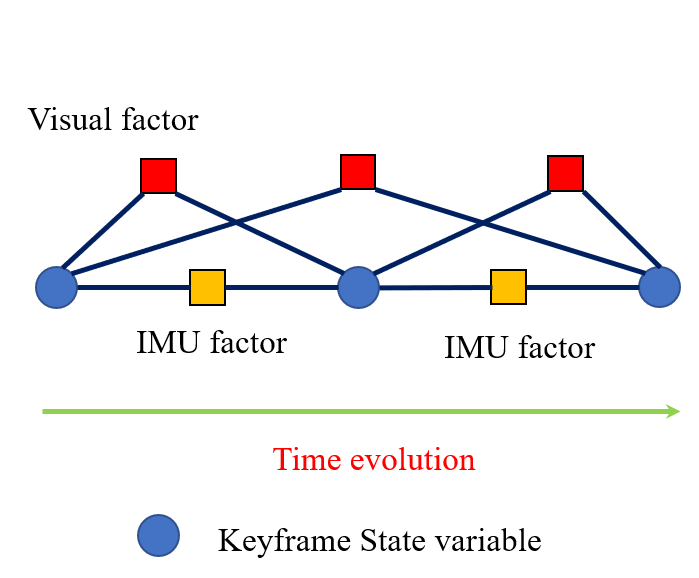}
        % \caption{Structureless VI-BA}
        \label{factor2}
    \end{subfigure}
  \caption{Left: factor graph for structure-based VI-BA. Right: factor graph for structureless VI-BA.}
  \label{factor}
\end{figure}

For visual constraints, most classical VIO systems inherit the design idea from VO, either using feature-based reprojection error \cite{mourikis2007multi, leutenegger2015keyframe, qin2018vins, geneva2020openvins, campos2021orb} or direct photometric error \cite{bloesch2017iterated, von2018direct}. The estimation of visual map (structure) and localization (pose) is interdependent. In order to remove the 3D positions or 1D depths of feature points from the state vector, the authors of Multi-State Constraint Kalman Filter (MSCKF) \cite{mourikis2007multi} proposed the idea of nullspace projection to modify the residual equation of the reprojection error and marginalize feature points. However, triangulation or depth estimation for feature points remains indispensable for the entire MSCKF system \cite{mourikis2007multi, geneva2020openvins}.
% This nullspace projection trick is also integrated into an optimization-based VIO \cite{demmel2021square, demmel2021square_iccv}.

To fully decouple the localization (pose) from visual map (structure), we design a novel structureless visual-inertial bundle adjustment (VI-BA) that naturally eliminates the dependence on the map (structure) by leveraging the epipolar constraint. The key difference between the classic structure-based VI-BA and the proposed structureless VI-BA is presented in Fig. \ref{factor}. By applying structureless VI-BA, this paper extends our previous work on monocular VIO initialization \cite{song2025improving} to subsequent sliding-window based optimization, yielding a novel and efficient structureless VIO.

\section{Structure-based VIO}

Before introducing the structureless VIO, we first briefly describe the structure-based VIO to better illustrate the differences in the context of keyframe-based VIO. Taking VINS-Mono \cite{qin2018vins} as an example, a structure-based VIO typically optimizes the following state in the sliding window
\begin{equation} \label{eq:sb}
    \begin{array}{l}
    x = {\left[ {\begin{array}{*{20}{c}}
    {x_{{c_0}}^T}& \cdots &{x_{{c_N}}^T}&{{\lambda}_{f_1}}& \cdots &{{\lambda}_{f_M}}
    \end{array}} \right]^T}\\
    x_{{c_k}} = {\left[ {\begin{array}{*{20}{c}}
    {{}^Gp_{{I_k}}^T}&{{}^Gv_{{I_k}}^T}&{{}_{{I_k}}^G{q^T}}&{b_{{a_k}}^T}&{b_{{g_k}}^T}
    \end{array}} \right]^T}
    \end{array}
\end{equation}

Where state vector $x$ includes $N + 1$ keyframe states and $M$ environmental features observed by these keyframes. ${x_{{c_k}}}$ represents the IMU state at keyframe timestamp ${t_k}$, including the IMU's position ${}^G{p_{{I_k}}}$, velocity ${}^G{v_{{I_k}}}$, orientation ${}_{{I_k}}^Gq$, accelerometer bias ${b_{{a_k}}}$, and gyroscope bias ${b_{{g_k}}}$. ${\lambda_{{f_l}}}$ denotes the inverse depth of an environmental feature point $f_l$.

The structure-based VIO can be formulated as the following structure-based VI-BA problem
\begin{equation} \label{structure-based VI-BA}
    \scalebox{0.9}{$
    \mathop {\min }\limits_x \left\{ {{\left\| {{r_{{p}}}} \right\|^2} + \sum\limits_{k = 1}^N {\left\| {{r_{{I_{k - 1,k}}}}} \right\|_{{\Sigma _{{I_{k - 1,k}}}}}^2}  + \sum\limits_{l = 1}^M {\sum\limits_{i \in {{\rm K}^l}} {{\rho _{Hub}}\left( {\left\| {{r_{il}}} \right\|_{{\Sigma _C}}^2} \right)} } } \right\}
    $}
\end{equation}

Where $r_{{p}}$, ${r_{{I_{k - 1,k}}}}\left( {{x_{{c_{k - 1}}}},{x_{{c_k}}}} \right)$ and ${r_{il}}$ are the prior residual from marginalization, IMU preintegration residual and visual reprojection residual, respectively. Their detailed definitions are provided in Section VI of \cite{qin2018vins}. ${K^l}$ is the set of keyframes observing feature point ${f_l}$. A robust Huber kernel function ${\rho _{Hub}}\left(  \bullet  \right)$ is used to mitigate the impact of pixel observation outliers.

\section{Proposed Structureless VIO}

We develop a novel structureless VIO based on VINS-Mono \cite{qin2018vins}. For feature tracking, IMU preintegration and sliding-window marginalization, we reuse the modules from VINS-Mono. And the key modification is the formulation of visual residual. The state variables of structureless VIO is obtained by deleting the environmental feature points
\begin{equation}
    \begin{array}{l}
    x = {\left[ {\begin{array}{*{20}{c}}
    {x_{{c_0}}^T}& \cdots &{x_{{c_N}}^T}
    \end{array}} \right]^T}\\
    x_{{c_k}} = {\left[ {\begin{array}{*{20}{c}}
    {{}^Gp_{{I_k}}^T}&{{}^Gv_{{I_k}}^T}&{{}_{{I_k}}^G{q^T}}&{b_{{a_k}}^T}&{b_{{g_k}}^T}
    \end{array}} \right]^T}
    \end{array}
\end{equation}

Unlike structure-based VIO, visual measurements are formulated by epipolar constraints, completely eliminating the dependence on 3D structure. Structureless VIO can be expressed as the following structureless VI-BA problem
\begin{equation} \label{eq:opt}
    \scalebox{0.85}{$
    \mathop {\min }\limits_x \left\{ {{\left\| {{r_{{p}}}} \right\|^2} + \sum\limits_{k = 1}^N {\left\| {{r_{{I_{k - 1,k}}}}} \right\|_{{\Sigma _{{I_{k - 1,k}}}}}^2}  + \sum\limits_{l = 1}^M {\sum\limits_{\left( {i,j} \right) \in {{\rm K}^l}} {{\rho _{Hub}}\left( {\left\| {r_{ij}^n} \right\|_{{\Sigma _C}}^2} \right)} } } \right\}
    $}
\end{equation}

Where $r_{{p}}$ and ${r_{{I_{k - 1,k}}}}\left( {{x_{{c_{k - 1}}}},{x_{{c_k}}}} \right)$ are described in Equation (\ref{structure-based VI-BA}). While $r_{ij}^n$ is the visual measurement residual generated by epipolar geometry. If a keyframe pair $\left( {i,j} \right)$ observes the same feature point ${f_l}$, this co-view relationship can be expressed using the epipolar constraint
\begin{equation}
    \begin{array}{l}
    r_{ij}^n\left( {{x_{{c_i}}},{x_{{c_j}}}} \right) = {\left( {{}_{{I_j}}^GR{}_C^IRz_j^n} \right)^T}{\left[ {\frac{t}{{\left\| t \right\|}}} \right]_ \times }\left( {{}_{{I_i}}^GR{}_C^IRz_i^n} \right)\\
    t = {}^G{p_{{C_i}}} - {}^G{p_{{C_j}}}\\
     = {}^G{p_{{I_i}}} + {}_{{I_i}}^GR{}^I{p_C} - {}^G{p_{{I_j}}} - {}_{{I_j}}^GR{}^I{p_C}
    \end{array}
\end{equation}

\begin{figure}[htbp]
    \centering
    \includegraphics[width=0.25\textwidth]{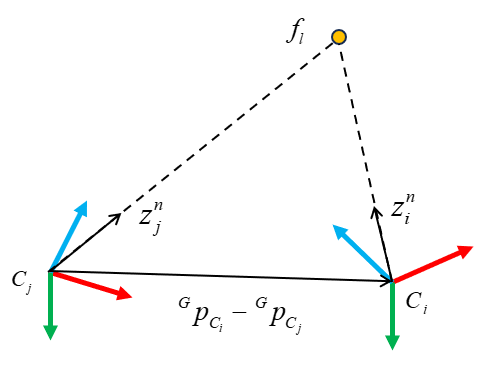}
    \caption{Co-planar geometric relationships for feature bearing vectors with the frame-to-frame translation vector.}
    \label{epipolar}
\end{figure}

Where $\left\{ {{}_C^IR,{}^I{p_C}} \right\}$ are the extrinsic parameters between IMU and camera. $z_i^n$ and $z_j^n$ represent the normalized coordinates observations of the same feature point ${f_l}$ from the keyframe pair $\left( {i,j} \right)$. $r_{ij}^n$ has intuitive geometric interpretation, i.e., two feature bearing vectors should be co-planar with the frame-to-frame translation vector $t$, as depicted in the Fig. \ref{epipolar}. All direction vectors are expressed in the global frame. $t$ is normalized to prevent it from converging to 0. Detailed Jacobians are provided in Section V-B of \cite{song2025improving}.

\section{Results}

Our structureless VIO is compared to the structure-based VINS-Mono \cite{qin2018vins}. The loop closure of VINS-Mono is disabled for fair comparison. For the setting of other parameters, we refer to VINS-Mono\footnote{\url{https://github.com/HKUST-Aerial-Robotics/VINS-Mono/tree/master/config}}. To verify the performance of different VIO algorithms, we employ two popular VIO datasets, EuRoC \cite{burri2016euroc} and TUM-VI \cite{schubert2018tum}. All the experiments are conducted on a laptop computer with an Intel(R) Xeon(R) W-10855M CPU @ 2.80GHz, and 16 GB of RAM.

Absolute trajectory error (ATE) \cite{zhang2018tutorial} and the average solve time are recorded in TABLE \ref{table_euroc} and TABLE \ref{table_tum}. Results demonstrate the structureless VIO scheme brings remarkable computational efficiency advantage over structure-based VIO. Moreover, structureless VIO can further improve accuracy and we attribute it to the independence of epipolar constraint from depth uncertainty, especially critical for corridor scenes.

\begin{table}
\caption{Performance comparison on EuRoC Dataset.}
\label{table_euroc}
\begin{center}
\begin{tabular}{|c|cc|cc|}
\hline
\multirow{2}{*}{\makecell{Sequence}} & \multicolumn{2}{c|}{ATE (m)} & \multicolumn{2}{c|}{Avg solve time (ms)}\\
\cline{2-5}   & VINS-Mono & Ours & VINS-Mono & Ours \\
\hline
MH\_01\_easy & \textbf{0.157} & 0.172 & 39.27  & \textbf{18.30} \\
MH\_02\_easy & \textbf{0.181} & 0.223 & 38.96  & \textbf{17.09} \\
MH\_03\_medium & \textbf{0.196} & 0.233 & 39.90  & \textbf{16.56} \\
MH\_04\_difficult & 0.378 & \textbf{0.257} & 37.76  & \textbf{15.57} \\
MH\_05\_difficult & 0.303 & \textbf{0.282} & 38.56  & \textbf{16.00} \\
\hline\hline
V1\_01\_easy & 0.082 & \textbf{0.050} & 39.22  & \textbf{16.44} \\
V1\_02\_medium & 0.112 & \textbf{0.084} & 33.54  & \textbf{13.27} \\
V1\_03\_difficult & 0.188 & \textbf{0.103} & 26.73  & \textbf{10.84} \\
V2\_01\_easy & 0.097 & \textbf{0.078} & 38.41  & \textbf{16.66} \\
V2\_02\_medium & 0.153 & \textbf{0.116} & 31.88  & \textbf{13.50} \\
V2\_03\_difficult & 0.298 & \textbf{0.245} & 20.87  & \textbf{10.40} \\
\hline\hline
\textcolor{red}{Avg} & 0.195 & \textbf{0.168} & 35.01 & \textbf{14.97} \\
\hline
\end{tabular}
\end{center}
\end{table}

\begin{table}
\caption{Performance comparison on TUM-VI Dataset.}
\label{table_tum}
\begin{center}
\begin{tabular}{|c|cc|cc|}
\hline
\multirow{2}{*}{\makecell{Sequence}} & \multicolumn{2}{c|}{ATE (m)} & \multicolumn{2}{c|}{Avg solve time (ms)}\\
\cline{2-5}   & VINS-Mono & Ours & VINS-Mono & Ours \\
\hline
room1 & 0.067 & \textbf{0.051} & 20.19  & \textbf{11.56} \\
room2 & \textbf{0.068} & 0.104 & 22.67  & \textbf{12.58} \\
room3 & 0.121 & \textbf{0.114} & 19.73  & \textbf{11.89} \\
room4 & \textbf{0.048} & 0.063 & 21.03  & \textbf{13.14} \\
room5 & 0.217 & \textbf{0.127} & 18.17  & \textbf{11.27} \\
room6 & \textbf{0.076} & 0.085 & 26.24  & \textbf{15.04} \\
\hline\hline
\textcolor{red}{Avg} & 0.100 & \textbf{0.091} & 21.34 & \textbf{12.58} \\
\hline\hline
corridor1 & 0.629 & \textbf{0.398} & 18.62  & \textbf{11.46} \\
corridor2 & \textbf{0.933} & 0.956 & 19.06  & \textbf{12.13} \\
corridor3 & 1.978 & \textbf{0.893} & 16.75  & \textbf{10.98} \\
corridor4 & 0.315 & \textbf{0.224} & 19.92  & \textbf{12.69} \\
corridor5 & 0.689 & \textbf{0.456} & 19.56  & \textbf{12.36} \\
\hline\hline
\textcolor{red}{Avg} & 0.909 & \textbf{0.585} & 18.78 & \textbf{11.92} \\
\hline
\end{tabular}
\end{center}
\end{table}

\section{Conclusion}

In this paper, we introduce a novel structureless VIO that adopts epipolar constraint instead of reprojection error or photometric error to formulate visual measurement. Experimental results on two publicly benchmark datasets demonstrate that our method reduces the optimization time by a large margin and offers improvement in accuracy.

% \addtolength{\textheight}{-12cm}   % This command serves to balance the column lengths
                                  % on the last page of the document manually. It shortens
                                  % the textheight of the last page by a suitable amount.
                                  % This command does not take effect until the next page
                                  % so it should come on the page before the last. Make
                                  % sure that you do not shorten the textheight too much.

%%%%%%%%%%%%%%%%%%%%%%%%%%%%%%%%%%%%%%%%%%%%%%%%%%%%%%%%%%%%%%%%%%%%%%%%%%%%%%%%

%%%%%%%%%%%%%%%%%%%%%%%%%%%%%%%%%%%%%%%%%%%%%%%%%%%%%%%%%%%%%%%%%%%%%%%%%%%%%%%%

%%%%%%%%%%%%%%%%%%%%%%%%%%%%%%%%%%%%%%%%%%%%%%%%%%%%%%%%%%%%%%%%%%%%%%%%%%%%%%%%
% \section*{APPENDIX}

% Appendixes should appear before the acknowledgment.

%\section*{ACKNOWLEDGMENT}

%This research was supported by the European Union’s Horizon 2020 project SESAME (grant agreement No 101017258). 

%%%%%%%%%%%%%%%%%%%%%%%%%%%%%%%%%%%%%%%%%%%%%%%%%%%%%%%%%%%%%%%%%%%%%%%%%%%%%%%%

\bibliographystyle{ieeetr}
\bibliography{bib}

\end{document}